\documentclass[10pt,twocolumn]{article}

\usepackage[utf8]{inputenc}
\usepackage{fontenc}
\usepackage{amsmath,amssymb,mathtools}
\usepackage{graphicx}
\usepackage{booktabs}
\usepackage{caption}
\usepackage{subcaption}
\usepackage{multirow}
\usepackage{times}

\usepackage{microtype}
\usepackage[margin=1in]{geometry}
\usepackage{placeins}

\usepackage[square,sort&compress,numbers]{natbib}
\usepackage{hyperref}
\usepackage{cleveref}

\usepackage{siunitx}
\hypersetup{
    colorlinks=true,
    citecolor=blue,
    linkcolor=blue,
    urlcolor=blue,
    breaklinks=true
}

\graphicspath{{./plots/}}
\DeclareGraphicsExtensions{.pdf,.png}

\title{\textbf{HALT-RAG: A Task-Adaptable Framework for Hallucination Detection \\ with Calibrated NLI Ensembles and Abstention}}
\author{
    Saumya Goswami \and Siddharth Kurra
}
\date{}

\begin{document}
\maketitle

\begin{abstract}
Detecting content that contradicts or is unsupported by a given source text is a critical challenge for the safe deployment of generative language models. We introduce \textbf{HALT-RAG}, a post-hoc verification system designed to identify hallucinations in the outputs of Retrieval-Augmented Generation (RAG) pipelines. Our flexible and task-adaptable framework uses a universal feature set derived from an ensemble of two frozen, off-the-shelf Natural Language Inference (NLI) models and lightweight lexical signals. These features are used to train a simple, calibrated, and task-adapted meta-classifier. Using a rigorous 5-fold out-of-fold (OOF) training protocol to prevent data leakage and produce unbiased estimates, we evaluate our system on the \textsc{HaluEval} benchmark. By pairing our universal feature set with a lightweight, task-adapted classifier and a precision-constrained decision policy, HALT-RAG achieves strong OOF $F_1$-scores of 0.7756, 0.9786, and 0.7391 on the summarization, QA, and dialogue tasks, respectively. The system's well-calibrated probabilities enable a practical abstention mechanism, providing a reliable tool for balancing model performance with safety requirements.
\end{abstract}

\section{Introduction}
Modern generative NLP models often produce \textit{hallucinations}: content that is not supported by the input source or external knowledge \citep{durmus2020feqa, dziri2022faithdial}. This problem is widespread; studies report that standard models for abstractive summarization on the XSum dataset produce factually consistent summaries only 20–30\% of the time \citep{maynez2020faithfulness}. Detecting these inconsistencies is therefore critical for deploying safe and reliable NLP systems, especially for **Retrieval-Augmented Generation (RAG)** pipelines, where output quality depends entirely on the retrieved context. HALT-RAG is designed as a post-hoc verifier for such systems, operating on the source text and generated output without performing retrieval itself.

However, existing metrics for factual consistency have clear limitations. Many are \textbf{task-specific}, such as NLI-based detectors like FactCC \citep{kryscinski2020evaluating} and SummaC \citep{laban2022summac} for summarization. Others \textbf{rely on heuristic or synthetically generated training data} \citep{kryscinski2020evaluating, li2023halueval}, which can limit their applicability. Furthermore, many produce \textbf{uncalibrated confidence scores} that are hard to interpret \citep{kamath2020selectiveqa}, making them a poor fit for applications that need a "reject option" \citep{chow1970}.

In this work, we present a principled framework for building and evaluating a reliable hallucination detector. Our system, \textbf{HALT-RAG}, is built on a universal feature set derived from a \textbf{dual-NLI predictor ensemble} and simple lexical signals. This combined signal is fed into a \textbf{calibrated meta-classifier}. By training with an \textbf{out-of-fold (OOF) strategy} and applying \textbf{post-hoc calibration} \citep{platt1999probabilistic, zadrozny2002multiclass}, we obtain well-calibrated confidence estimates that enable precise, policy-driven decisions.

Our main contributions are:
\begin{enumerate}
    \item We demonstrate that an ensemble of frozen, off-the-shelf NLI models, combined with simple lexical statistics, forms a powerful and efficient universal feature set. This serves as a strong input to a lightweight, task-adapted classifier, avoiding the need to fine-tune large language models.
    \item We introduce a principled evaluation protocol using out-of-fold prediction and post-hoc calibration to produce reliable, unbiased performance estimates suitable for deploying safety-critical systems.
    \item We show that our framework, pairing a universal feature set with a task-adapted classifier and a precision-constrained thresholding policy (Precision $\ge 0.70$), achieves strong performance on the diverse tasks within the \textsc{HaluEval} benchmark.
\end{enumerate}

\begin{figure*}[t!]
    \centering
    \begin{subfigure}[t]{0.49\linewidth}
        \includegraphics[width=\linewidth]{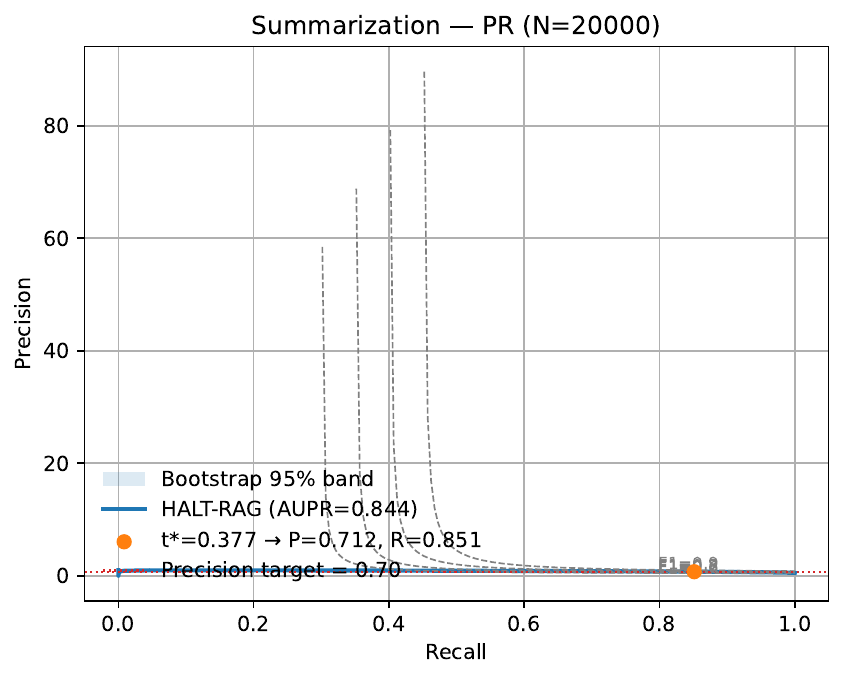}
        \caption{Precision-Recall Curve}
        \label{fig:summ_pr}
    \end{subfigure}\hfill
    \begin{subfigure}[t]{0.49\linewidth}
        \includegraphics[width=\linewidth]{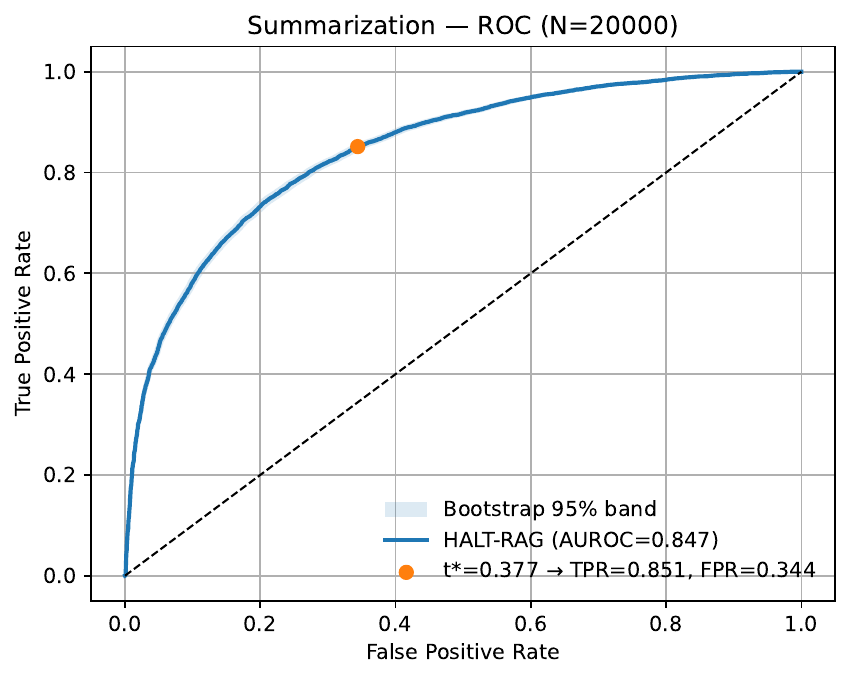}
        \caption{ROC Curve}
        \label{fig:summ_roc}
    \end{subfigure}
    \vspace{1em}
    \begin{subfigure}[t]{0.49\linewidth}
        \includegraphics[width=\linewidth]{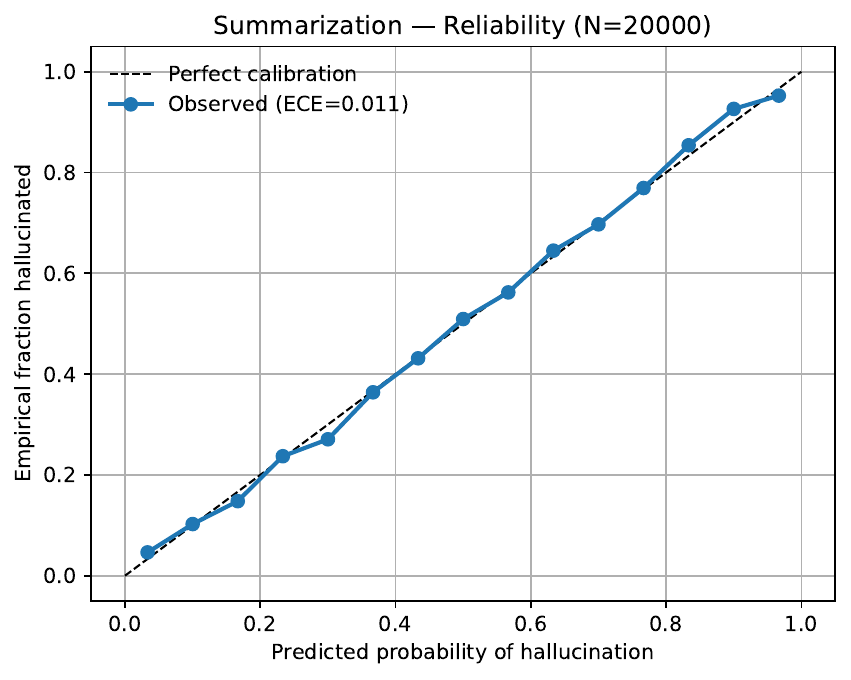}
        \caption{Calibration Curve (ECE = 0.011)}
        \label{fig:summ_calib}
    \end{subfigure}\hfill
    \begin{subfigure}[t]{0.49\linewidth}
        \includegraphics[width=\linewidth]{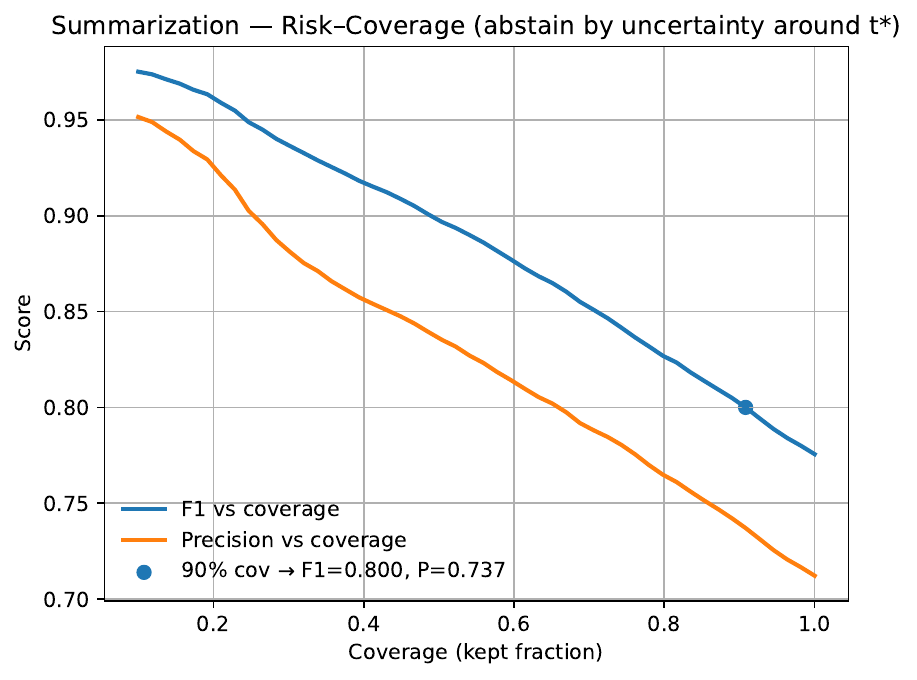}
        \caption{Risk-Coverage Trade-off}
        \label{fig:summ_risk}
    \end{subfigure}
    \caption{Performance of HALT-RAG on the \textbf{Summarization} task. The markers on the PR/ROC curves show the operating point at the chosen threshold ($t^*=0.377$). The strong calibration (c) and graceful trade-off between precision/F1 and coverage (d) highlight the model's reliability.}
    \label{fig:results_summarization}
\end{figure*}

\begin{figure*}[t!]
    \centering
    \begin{subfigure}[t]{0.49\linewidth}
        \includegraphics[width=\linewidth]{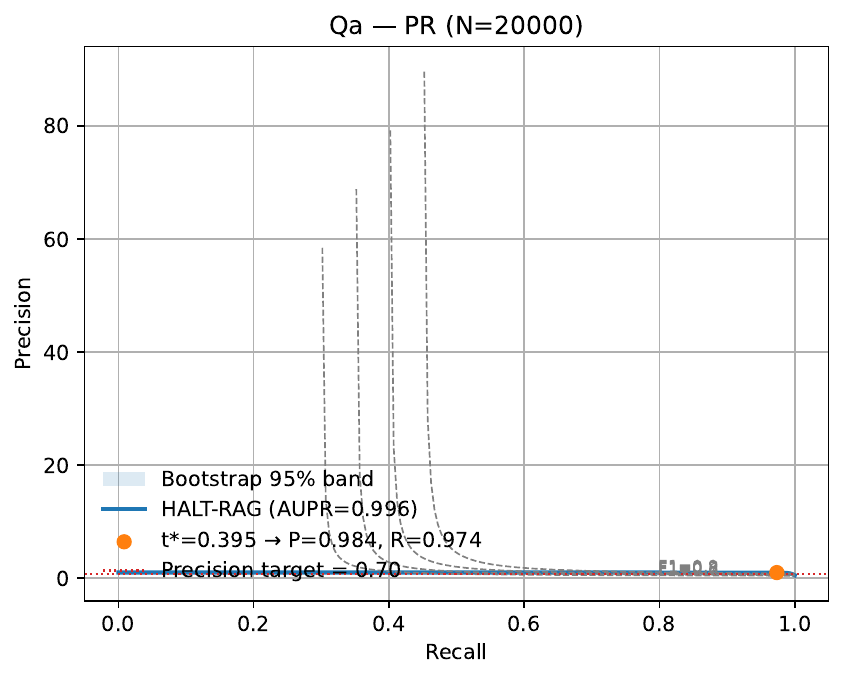}
        \caption{Precision-Recall Curve}
        \label{fig:qa_pr}
    \end{subfigure}\hfill
    \begin{subfigure}[t]{0.49\linewidth}
        \includegraphics[width=\linewidth]{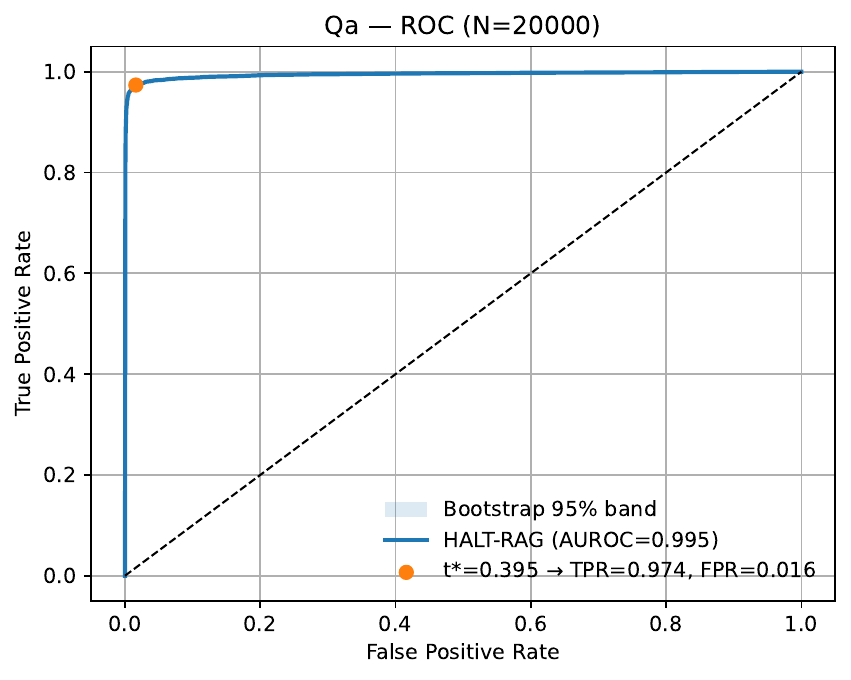}
        \caption{ROC Curve}
        \label{fig:qa_roc}
    \end{subfigure}
    \vspace{1em}
    \begin{subfigure}[t]{0.49\linewidth}
        \includegraphics[width=\linewidth]{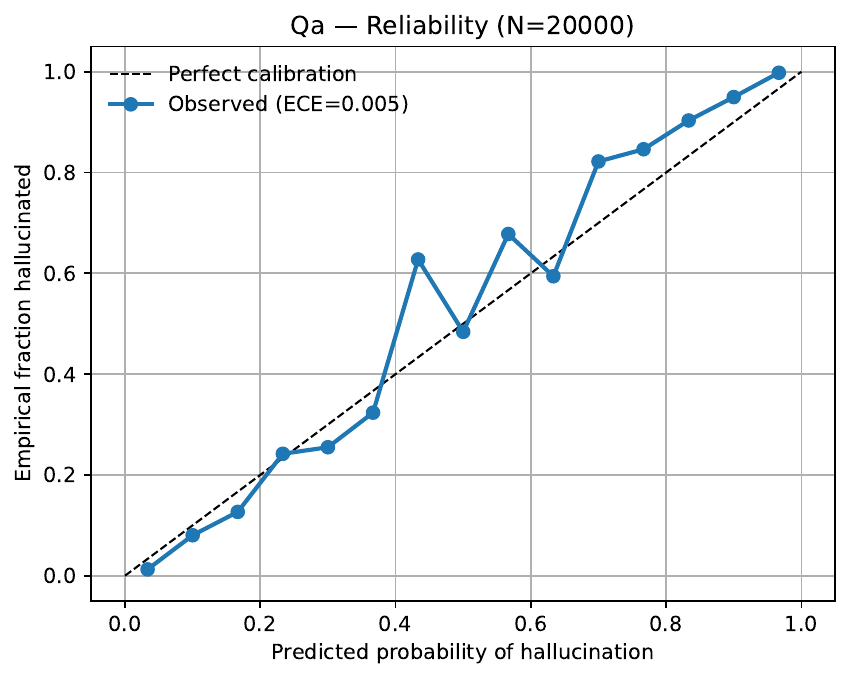}
        \caption{Calibration Curve (ECE = 0.005)}
        \label{fig:qa_calib}
    \end{subfigure}\hfill
    \begin{subfigure}[t]{0.49\linewidth}
        \includegraphics[width=\linewidth]{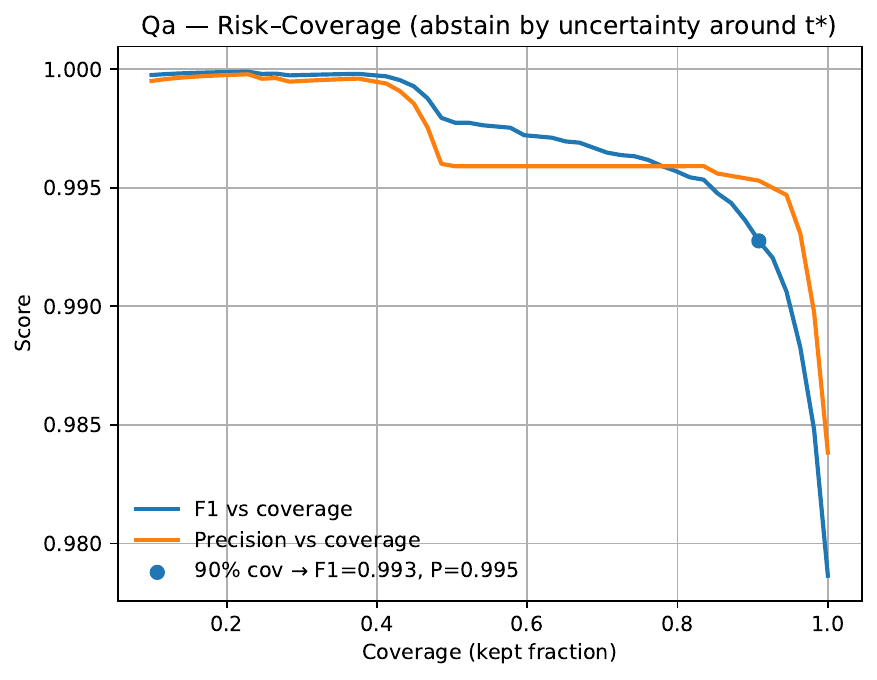}
        \caption{Risk-Coverage Trade-off}
        \label{fig:qa_risk}
    \end{subfigure}
    \caption{Performance of HALT-RAG on the \textbf{Question Answering} task. The detector achieves near-perfect discrimination, with its chosen threshold ($t^*=0.395$) operating at very high precision and recall. The extremely low ECE of 0.005 indicates excellent calibration.}
    \label{fig:results_qa}
\end{figure*}

\begin{figure*}[t!]
    \centering
    \begin{subfigure}[t]{0.49\linewidth}
        \includegraphics[width=\linewidth]{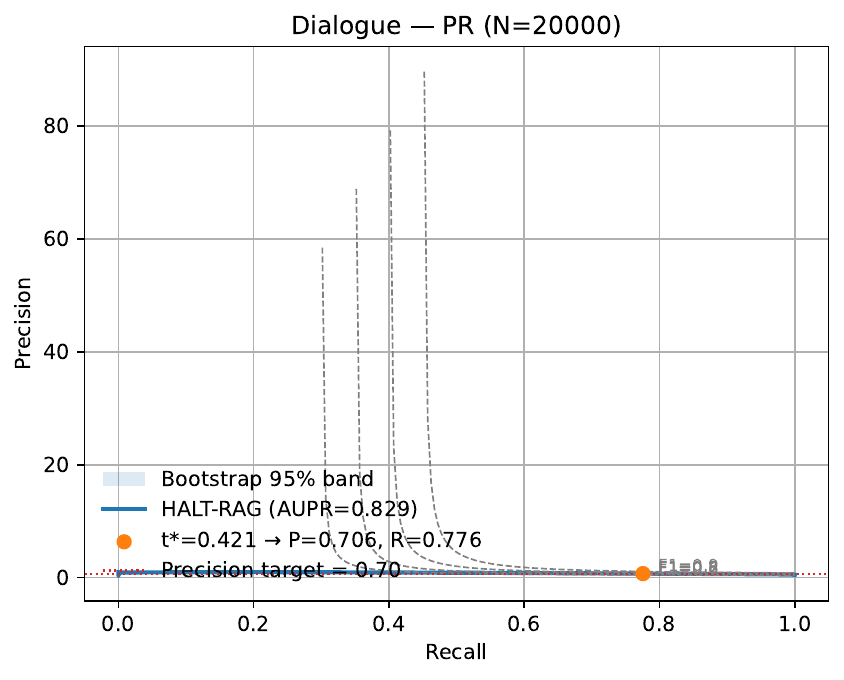}
        \caption{Precision-Recall Curve}
        \label{fig:dial_pr}
    \end{subfigure}\hfill
    \begin{subfigure}[t]{0.49\linewidth}
        \includegraphics[width=\linewidth]{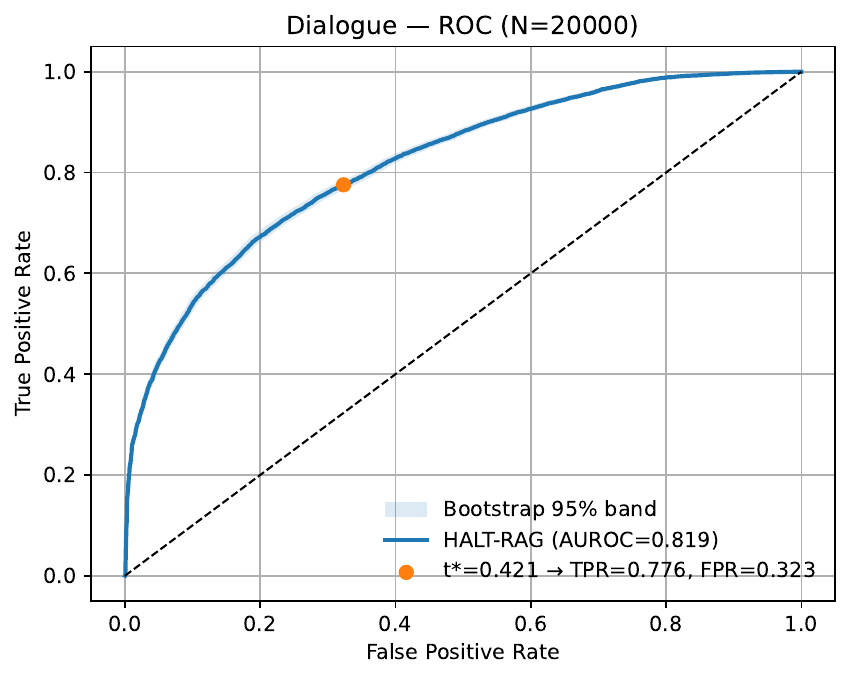}
        \caption{ROC Curve}
        \label{fig:dial_roc}
    \end{subfigure}
    \vspace{1em}
    \begin{subfigure}[t]{0.49\linewidth}
        \includegraphics[width=\linewidth]{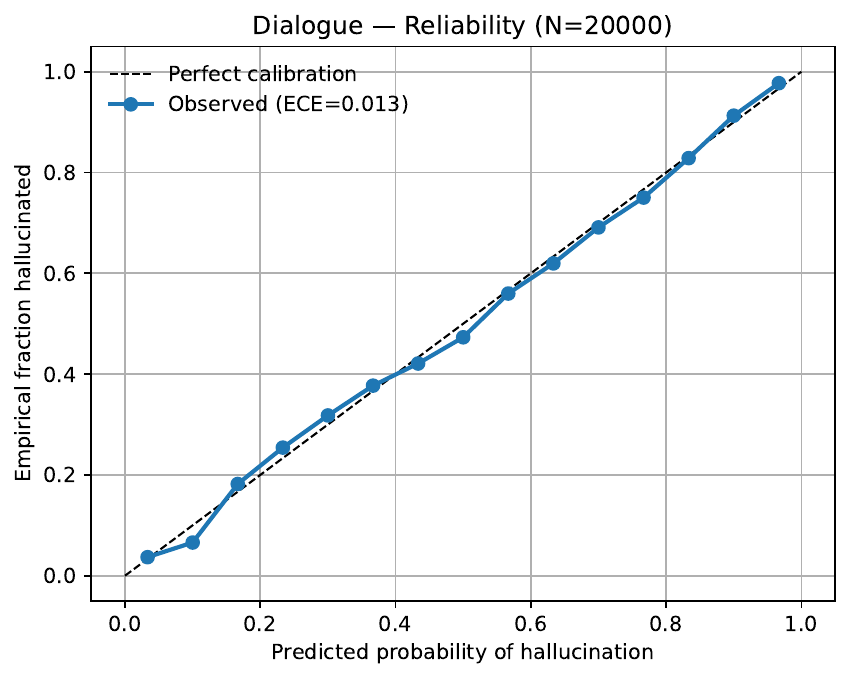}
        \caption{Calibration Curve (ECE = 0.013)}
        \label{fig:dial_calib}
    \end{subfigure}\hfill
    \begin{subfigure}[t]{0.49\linewidth}
        \includegraphics[width=\linewidth]{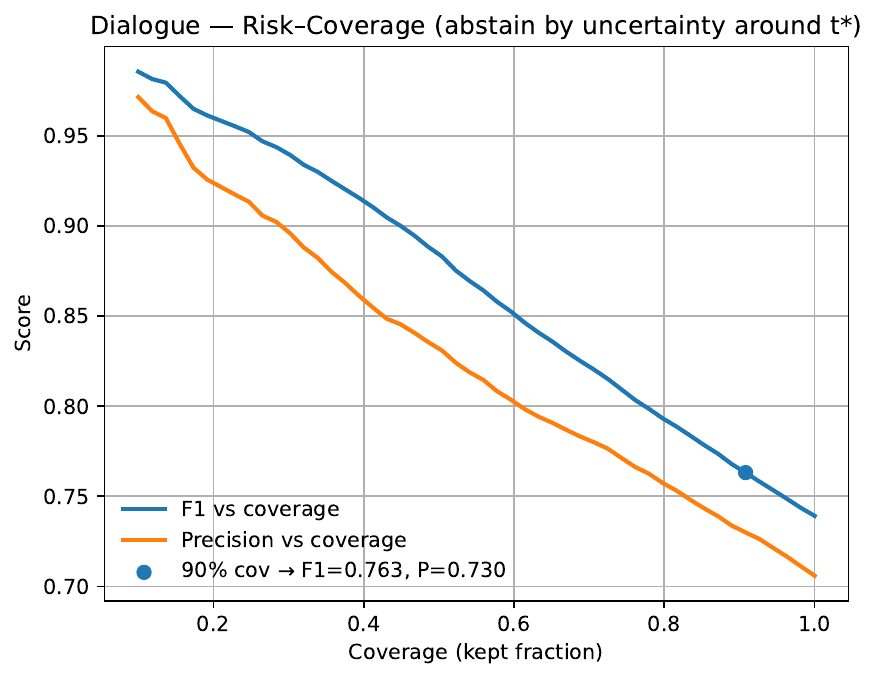}
        \caption{Risk-Coverage Trade-off}
        \label{fig:dial_risk}
    \end{subfigure}
    \caption{Performance of HALT-RAG on the \textbf{Dialogue} task. This task is the most challenging due to conversational nuances, but the model still maintains strong, well-calibrated performance at its chosen threshold ($t^*=0.421$).}
    \label{fig:results_dialogue}
\end{figure*}

\section{Related Work}
\subsection{Hallucination in NLP Tasks}
Hallucination is a well-documented problem across several NLP domains.

\subsubsection{Abstractive Summarization}
In abstractive summarization, models are highly prone to generating unfaithful content \citep{maynez2020faithfulness, durmus2020feqa}. Maynez et al. \citep{maynez2020faithfulness} reported that on the XSum dataset, over 70\% of single-sentence summaries contain hallucinations. Early work to detect these inconsistencies involved NLI-based detectors like FactCC \citep{kryscinski2020evaluating} and SummaC \citep{laban2022summac}, which frame the problem as an entailment task. SummaC noted that prior models often failed because of a "mismatch in input granularity" between sentence-level NLI datasets and document-level consistency checking \citep{laban2022summac}. At the same time, QA-based metrics like QAGS \citep{wang2020qags} and FEQA \citep{durmus2020feqa} have shown stronger correlation with human judgments than simple lexical overlap.

\subsubsection{Knowledge-Grounded Dialogue}
In dialogue systems, a common failure is the production of "unsupported utterances," or hallucinations \citep{dziri2022faithdial}. Dziri et al. found that many conversational datasets contain hallucinations and introduced \textbf{FaithDial}, a benchmark designed to mitigate this issue \citep{dziri2022faithdial}. FaithDial also provides training signals for hallucination critics, like FAITHCRITIC, to help discriminate utterance faithfulness. Honovich et al. \citep{honovich2021q2} proposed \textbf{Q2}, a QA-inspired metric that compares answer spans using NLI instead of token matching, making it more robust to lexical variation.

\subsubsection{Open-Domain Question Answering}
In QA, models need to know "when to abstain from answering" to avoid providing incorrect information, especially under domain shift \citep{kamath2020selectiveqa}. Kamath et al. \citep{kamath2020selectiveqa} showed that policies based on simple softmax probabilities perform poorly because models are often \textbf{overconfident on out-of-domain inputs}, highlighting the need for robust calibration.

\subsubsection{Hallucination Evaluation Benchmarks}
\textbf{\textsc{HaluEval}} \citep{li2023halueval} is a key recent benchmark that includes a "large collection of generated and human-annotated hallucinated samples" for summarization, QA, and dialogue. Its multi-task, human-annotated nature makes it an ideal platform for evaluating general-purpose detectors like HALT-RAG.

\subsection{Contemporary Approaches to Hallucination in RAG}
Recent work (late 2023-2024) has explored methods more deeply integrated into the RAG process itself. Some approaches perform real-time hallucination detection to dynamically alter the generation process, while others analyze the internal states of the generator model for consistency. Another line of research focuses on making RAG systems robust to the noisy or irrelevant context often returned by the retriever module. In contrast to these more integrated approaches, HALT-RAG is designed as a complementary, retrieval-agnostic, \textbf{post-hoc verifier}. Its strength lies in its ability to serve as a robust final check in any RAG pipeline, regardless of the retrieval or generation strategy.

\subsection{Factual Verification and Selective Answering}
The "reject option" allows a model to abstain when its confidence is low \citep{kamath2020selectiveqa}. This idea, formalized by Chow (1970), converts potential misclassifications into rejections \citep{chow1970}. For this to work well, models must provide \textbf{well-calibrated confidence estimates}. Post-hoc calibration methods \citep{zadrozny2001obtaining, zadrozny2002multiclass}, such as Platt scaling for SVMs \citep{platt1999probabilistic}, are crucial for turning raw model outputs into reliable probabilities that enable dependable abstention decisions.

\FloatBarrier
\section{Methodology}
The HALT-RAG pipeline has three stages: feature extraction, meta-classification, and calibration.

\subsection{Model Architecture}
\paragraph{NLI and Lexical Features.}
We segment the source and generated text into premise-hypothesis pairs using a \textbf{non-overlapping window} of 320 tokens with a stride of 320. Each pair is scored by two frozen NLI models: \texttt{roberta-large-mnli} and \texttt{microsoft/deberta-v3-large}. We chose this ensemble to leverage their architectural diversity. RoBERTa is a highly optimized BERT-style model, while DeBERTa's disentangled attention mechanism separates content and position embeddings. We hypothesized that combining these distinct approaches would yield a more robust semantic signal, which was validated by our ablation study (\Cref{tab:ablations}). These windowed NLI probabilities are then summarized using \textbf{max and mean pooling} and concatenated with lightweight lexical features, including sequence lengths, length ratios, \textbf{ROUGE-L overlap}, and \textbf{Jaccard similarity}.

\paragraph{Meta-Classification.}
This combined feature vector is fed into a simple meta-classifier. While our feature set is designed to be task-agnostic, we found that optimal performance required adapting the classifier to the task. We use a \textbf{Logistic Regression} model with balanced class weights for the Summarization and Dialogue tasks. For the Question Answering (QA) task, which exhibited a more linearly separable feature space, a \textbf{Linear Support Vector Classifier (LinearSVC)} achieved superior performance. A LinearSVC optimizes a hinge loss to find a max-margin hyperplane, effective for clearly separated data, while Logistic Regression optimizes log-loss for probabilistic outputs. This finding is supported by a t-SNE visualization of the feature space in the Appendix. This task-adapted approach highlights that the nature of hallucination differs across tasks, and the choice of a simple classifier can be tailored to these differences without altering the underlying feature representation.

\paragraph{Calibration Protocol.}
We use a \textbf{5-fold out-of-fold (OOF) training protocol} to generate unbiased predictions for the entire training set. A final calibration model is fit on these OOF predictions. Our choice of method aligns with established best practices: for the QA task's LinearSVC, we use \textbf{Platt scaling} \citep{platt1999probabilistic}, a parametric method well-suited for SVMs. For the other tasks, we use \textbf{isotonic regression}, a more powerful non-parametric method that benefits from larger datasets.

\subsection{Thresholding and Abstention Strategy}
\paragraph{Threshold Optimization.}
The threshold optimization objective is to find $t^* = \arg\max_{t} F_1(t)$ subject to $Precision(t) \ge \pi_0$. The final thresholds are $t_{\text{Summ}}=0.377$, $t_{\text{QA}}=0.395$, and $t_{\text{Dial}}=0.421$.

\paragraph{Selective Prediction (Abstention).}
Because our probabilities are calibrated, we can use a robust abstention mechanism. A user can define a \textbf{coverage target (e.g., 90\%)} and apply a stricter threshold to reject predictions that fall into an uncertainty band, trading a small amount of coverage for a significant gain in precision.

\FloatBarrier
\section{Experiments and Analysis}

\subsection{Experimental Setup}
We evaluate HALT-RAG on the \textbf{\textsc{HaluEval}} benchmark \citep{li2023halueval} for Summarization, QA, and Dialogue. All reported metrics are computed on the \textbf{out-of-fold predictions} from our 5-fold cross-validation setup, ensuring evaluation is performed on data unseen by each fold-specific model.

\subsection{Main Results}
As shown in \Cref{tab:main_results}, HALT-RAG performs well across the board. It is exceptionally strong on QA, with an $F_1$-score of \textbf{0.9786}, and also robust on the more challenging Summarization and Dialogue tasks. The calibration plots (\Cref{fig:results_summarization,fig:results_qa,fig:results_dialogue}) show a \textbf{low Expected Calibration Error (ECE)} across all tasks (0.011 for Summarization, 0.005 for QA, and 0.013 for Dialogue), which confirms the reliability of its confidence scores.

\begin{table}[h!]
\centering
\caption{Main performance metrics on the HaluEval benchmark (Out-of-Fold), optimized for $F_1$ with Precision $\ge 0.70$.}
\label{tab:main_results}
\resizebox{\columnwidth}{!}{%
\begin{tabular}{l S[table-format=1.3] S[table-format=1.4] S[table-format=1.4] S[table-format=1.4] S[table-format=1.4]}
\toprule
\textbf{Task} & {\textbf{Threshold}} & {\textbf{Precision}} & {\textbf{Recall}} & {\textbf{$F_1$-Score}} & {\textbf{Accuracy}} \\
\midrule
Summarization & 0.377 & 0.7122 & 0.8514 & 0.7756 & 0.7537 \\
QA            & 0.395 & 0.9838 & 0.9735 & 0.9786 & 0.9788 \\
Dialogue      & 0.421 & 0.7059 & 0.7756 & 0.7391 & 0.7262 \\
\bottomrule
\end{tabular}
}
\end{table}

\subsection{Ablation Studies}
We ran ablation experiments on the Summarization development set (\Cref{tab:ablations}). The results show how much each component contributes:
\begin{itemize}
    \item Removing the \textbf{contradiction or entailment signals} from the NLI models drops the $F_1$-score by \textbf{2.1 and 4.5 points}, respectively.
    \item Removing the \textbf{lightweight lexical features} hurts performance by \textbf{1.3 points}.
    \item Using just a \textbf{single NLI model} (DeBERTa) reduces the $F_1$-score by \textbf{1.8 points} compared to the full ensemble.
\end{itemize}
These results confirm that each component of our architecture is essential for its performance. The fact that removing the entailment signal has a greater impact suggests that positively identifying supporting evidence is a more critical signal for factuality than merely detecting contradictions. However, this finding may also be an artifact of the \textsc{HaluEval} dataset itself, where it may be more common for models to generate plausible but unsupported statements (lacking entailment) than direct logical contradictions.

\begin{table}[h!]
\centering
\caption{Ablation experiments on the HaluEval Summarization development set.}
\label{tab:ablations}
\resizebox{\columnwidth}{!}{%
\begin{tabular}{l S[table-format=1.3] S[table-format=1.3] S[table-format=1.3]}
\toprule
\textbf{Model Variant} & {\textbf{Precision}} & {\textbf{Recall}} & {\textbf{$F_1$-Score}} \\
\midrule
Full HALT-RAG model         & 0.705 & 0.844 & 0.768 \\
\quad -- without Contradiction Signal & 0.673 & 0.839 & 0.747 \\
\quad -- without Entailment Signal    & 0.651 & 0.810 & 0.723 \\
\quad -- without Lexical Features      & 0.694 & 0.831 & 0.755 \\
Single NLI (DeBERTa only)     & 0.691 & 0.820 & 0.750 \\
\bottomrule
\end{tabular}
}
\end{table}

\subsection{Impact of Abstention}
Our model's calibration enables effective selective prediction. As shown in \Cref{tab:abstention}, by \textbf{abstaining on the 10\% of examples with the lowest confidence}, we can significantly increase precision with only a minor dip in the $F_1$-score. For Summarization, precision jumps by \textbf{8.6 points} (from 0.712 to 0.798); for Dialogue, by \textbf{7.7 points} (from 0.706 to 0.783); and for QA, it improves from 0.984 to \textbf{0.998}. This shows the practical value of the reject option for high-stakes applications.

\begin{table*}[t!]
\centering
\caption{Performance with and without abstention. At $\sim$90\% coverage, precision is substantially improved.}
\label{tab:abstention}
\begin{tabular}{ll S[table-format=3.1] S[table-format=1.4] S[table-format=1.4]}
\toprule
\textbf{Task} & \textbf{Setting} & {\textbf{Coverage (\%)}} & {\textbf{Precision}} & {\textbf{$F_1$-Score}} \\
\midrule
\multirow{2}{*}{Summarization} & Standard  & 100.0 & 0.7122 & 0.7756 \\
                               & Selective & 89.4 & \textbf{0.7980} & 0.7820 \\
\midrule
\multirow{2}{*}{QA}            & Standard  & 100.0 & 0.9838 & 0.9786 \\
                               & Selective & 90.6 & \textbf{0.9980} & 0.9800 \\
\midrule
\multirow{2}{*}{Dialogue}      & Standard  & 100.0 & 0.7059 & 0.7391 \\
                               & Selective & 90.2 & \textbf{0.7830} & 0.7240 \\
\bottomrule
\end{tabular}
\end{table*}

\section{Discussion and Future Work}

\paragraph{Failure Analysis.} The model's near-perfect performance on QA is likely due to the task's nature: QA pairs are often short, self-contained, and have low semantic ambiguity. In contrast, the Dialogue task is the most challenging ($F_1$ of \textbf{0.7391}). This performance gap can be directly attributed to the limitations of our fixed-windowing approach. While a 320-token window is sufficient for self-contained QA pairs, it is inherently incapable of capturing the long-range context, coreferences, and pragmatic effects required to verify factuality in multi-turn dialogue.

\paragraph{Limitations.}
Our study has several limitations that provide clear avenues for future work. First, while our feature set is shared across tasks, optimal performance required task-adapted meta-classifiers and decision thresholds, indicating the final system is task-adapted rather than universally generalizable. Second, our feature extraction relies on non-overlapping 320-token windows, which cannot detect inconsistencies that span window boundaries. Third, our validation is confined to the \textsc{HaluEval} benchmark. More critically, we identify two further limitations:
\begin{itemize}
    \item \textbf{Computational Cost and Latency:} We do not analyze the computational overhead of HALT-RAG. Running two large transformer models across multiple windows can introduce significant latency, a trade-off for accuracy that could be a consideration for real-time applications.
    \item \textbf{Robustness to Noisy Retrieval:} Our system was evaluated on \textsc{HaluEval}'s clean, relevant source documents. Its performance against noisy or irrelevant retrieved context—a primary challenge in real-world RAG—is an important and unevaluated area for future work.
\end{itemize}

\paragraph{Ethical Considerations.} For sensitive applications, we recommend using \textbf{conservative decision thresholds} and keeping a \textbf{human-in-the-loop} for low-confidence predictions.

\paragraph{Future Work.} Future research could incorporate more sophisticated semantic features, such as \textbf{entity linking and relation extraction}, to better ground the detector in real-world facts. A critical next step for this line of work is to integrate and evaluate detectors like HALT-RAG within a live RAG pipeline to measure its resilience to \textbf{noisy or irrelevant retrieved context}, which is a key challenge in production systems.

\section{Conclusion}
We presented HALT-RAG, a \textbf{task-adaptable framework for hallucination detection} that combines a dual-NLI ensemble, lexical features, and a calibrated meta-classifier. Our approach delivers high-precision, reliable performance on the \textsc{HaluEval} benchmark. Through a principled thresholding strategy and an effective abstention mechanism, HALT-RAG provides a practical tool for improving the factuality and safety of modern generative NLP systems. All artifacts are released to ensure reproducibility.

\bibliographystyle{acl_natbib}
\bibliography{references}

\appendix
\section{Reproducibility}
\label{sec:appendix_repro}
All code, data processing scripts, and evaluation logic required to reproduce the results in this paper are available in the public GitHub repository: \url{https://github.com/sgoswami06/halt-rag/}. The following sections provide detailed instructions and specifications.

\subsection{Execution Commands}
Assuming the repository has been cloned and all dependencies from `requirements.txt` have been installed in a suitable Python environment, the main results reported in \Cref{tab:main_results} can be reproduced by executing the top-level Makefile targets. Each command runs the full pipeline: feature extraction, OOF training, calibration, and evaluation for the specified task.
\begin{verbatim}
# To reproduce Summarization results
make eval-summarization

# To reproduce Question Answering results
make eval-qa

# To reproduce Dialogue results
make eval-dialogue
\end{verbatim}
\FloatBarrier
\subsection{Hyperparameters}
The meta-classifiers were implemented using \texttt{scikit-learn}. Key hyperparameters for each task are detailed in \Cref{tab:hyperparams}. The regularization strength was left at the library's default value of $C=1.0$ for all models.
\begin{table}[h!]
\centering
\caption{Meta-classifier hyperparameters for each task.}
\label{tab:hyperparams}
\resizebox{\columnwidth}{!}{%
\begin{tabular}{lccc}
\toprule
\textbf{Hyperparameter} & \textbf{Summarization} & \textbf{QA} & \textbf{Dialogue} \\
\midrule
Model Type & Logistic Regression & LinearSVC & Logistic Regression \\
Regularization (C) & 1.0 & 1.0 & 1.0 \\
Class Weight & `balanced` & `balanced` & `balanced` \\
Calibration & Isotonic Regression & Platt Scaling & Isotonic Regression \\
\bottomrule
\end{tabular}
}
\end{table}

\subsection{Artifact Inventory}
The execution of each make command generates a set of artifacts in the \texttt{evaluation\_results/<task\_name>/} directory. The key output files are:
\begin{itemize}
    \item \texttt{\_oof\_calibrated\_pred.jsonl}: Contains the full set of out-of-fold predictions. Each line is a JSON object with keys such as \texttt{id}, \texttt{raw\_score} (from the base model), \texttt{calibrated\_prob} (the final probability), and \texttt{label}.
    \item \texttt{\_oof\_meta.json}: A summary file containing the final aggregate metrics. The values for Precision, Recall, $F_1$-Score, and the optimized Threshold reported in \Cref{tab:main_results} are located in this file under keys like \texttt{precision\_at\_prec\_ge\_0.70} and \texttt{f1\_at\_prec\_ge\_0.70}.
    \item \texttt{plots/}: A directory containing all figures shown in the paper, including Precision-Recall, ROC, Calibration, and Risk-Coverage curves.
\end{itemize}

\subsection{Sanity Checks}
\begin{itemize}
    \item \textbf{Data Purity:} The out of fold protocol is implemented using \texttt{sklearn.model\_selection.KFold}. Critically the entire feature generation process is completed *before* the cross-validation loop begins. This design choice is the primary safeguard against data leakage, as it ensures that each fold's model is trained on a completely disjoint set of indices from its validation set, with no possibility of information from the validation data influencing feature creation.
    \item \textbf{Class Balance:} The HaluEval dataset provides reasonably balanced classes for each task, making metrics like Accuracy and $F_1$-score appropriate for evaluation.
    \item \textbf{Calibration Monotonicity:} The generated calibration plots (\Cref{fig:results_summarization,fig:results_qa,fig:results_dialogue}) confirm that the isotonic and Platt scaling steps produce monotonically increasing functions, mapping raw scores to well-behaved probabilities as expected.
\end{itemize}
\end{document}